\documentclass[letterpaper, 10 pt, conference]{IEEEtran}  

\IEEEoverridecommandlockouts                              

\usepackage{amsmath} 
\usepackage{amssymb}  
\usepackage{mathtools}
\usepackage{booktabs}
\usepackage{graphicx}
\usepackage{float}
\usepackage{subfigure}

\usepackage{background}
\usepackage{stackengine}

\title{\LARGE \bf
Passive and Active Learning of Driver Behavior from Electric Vehicles
}

\author{Federica Comuni$^{1,\star}$, Christopher Mészáros$^{2,\star}$, Niklas Åkerblom$^{3}$ and Morteza Haghir Chehreghani$^{4}$
\thanks{$^{\star}$Equal contribution}
\thanks{$^{1}$Federica Comuni is with the Department of CSE, University of Gothenburg, Sweden
        {\tt\small federica.comuni@gmail.com}}%
\thanks{$^{2}$Christopher Mészáros is with the Department of CSE, Chalmers University of Technology, Sweden
        {\tt\small christophermeszaros07@gmail.com}}%
\thanks{$^{3}$Niklas Åkerblom is with the Department of CSE, Chalmers University of Technology, Sweden, and with Volvo Car Corporation, Sweden
        {\tt\small niklas.akerblom@chalmers.se}}%
\thanks{$^{4}$Morteza Haghir Chehreghani is with the Department of CSE, Chalmers University of Technology, Sweden
        {\tt\small morteza.chehreghani@chalmers.se}}%
}

\begin{document}

\maketitle
\thispagestyle{empty}
\pagestyle{empty}

\setstackEOL{\\}
\SetBgContents{\stackunder[37.5cm]{\Longstack{ This article has been accepted for publication in IEEE 25th International Conference on Intelligent Transportation Systems (ITSC). This is the author's version which\\
has not been fully edited and content may change prior to final publication. Citation information: DOI 10.1109/ITSC55140.2022.9922012}}{$\copyright$ 2022 IEEE. Personal use is permitted, but republication/redistribution requires IEEE permission. See https://www.ieee.org/publications/rights/index.html for more information.}}
\SetBgScale{0.7}
\SetBgAngle{0}
\SetBgPosition{current page.center}
\SetBgVshift{0.5cm}
\SetBgColor{black}
\SetBgOpacity{1}

\begin{abstract}
Modeling driver behavior provides several advantages in the automotive industry, including prediction of electric vehicle energy consumption. Studies have shown that aggressive driving can consume up to $30\%$ more energy than moderate driving, in certain driving scenarios. Machine learning methods are widely used for driver behavior classification, which, however, may yield some challenges such as sequence modeling on long time windows and lack of labeled data due to expensive annotation. 

To address the first challenge, passive learning of driver behavior, we investigate non-recurrent architectures such as self-attention models and convolutional neural networks with joint recurrence plots (JRP), and compare them with recurrent models. We find that self-attention models yield good performance, while JRP does not exhibit any significant improvement. However, with the window lengths of 5 and 10 seconds used in our study, none of the non-recurrent models outperform the recurrent models.

To address the second challenge, we investigate several active learning methods with different informativeness measures. We evaluate uncertainty sampling, as well as more advanced methods, such as query by committee and active deep dropout. Our experiments demonstrate that some active sampling techniques can outperform random sampling, and therefore decrease the effort needed for annotation. 

\end{abstract}

\section{INTRODUCTION}

Several applications in the automotive industry can benefit from a driver behavior model. Examples of such applications include advanced driver-assistance systems (ADAS) and estimation of vehicle energy consumption. ADAS can monitor and alert the driver when potentially dangerous behaviors are displayed \cite{Lin2014a}. 
A driver behavior model can help ADAS to better identify and understand driver characteristics that are unsafe or aggressive. 

Prediction of vehicle energy consumption of upcoming trips can also benefit from the ability to detect driver behaviors. Several studies have shown that the behavior of the driver significantly affects the energy consumption of electric vehicles. More specifically, aggressive driving can consume up to 30\% more energy than moderate driving \cite{Bingham2012a}. Therefore, identification of aggressive driving can yield a more precise estimation of the energy consumption and the remaining range of a vehicle, which can further be used for better navigation functionality \cite{Akerblom0C20,abs-2111-02314}.  Creating an effective and useful driver behavior model, however, requires overcoming some important challenges. In this paper we investigate and address two of these challenges.

The first challenge arises when modeling driver behavior using artificial neural networks with a recurrent architecture, such as the \textit{gated recurrent unit} (GRU) or the \textit{long short-term memory} (LSTM), an approach often preferred due to the time-series nature of vehicle sensor data (e.g., \cite{Ping2019b} or \cite{Xing2020a}). These types of networks are designed to be sequentially fed with data. Since the hardware (i.e., GPUs) typically used to train neural networks leverages parallel computations for efficiency, this can result in relatively long training times, especially with input data consisting of longer time series. Hence, there is a demand for alternative architectures capable of utilizing better the properties of GPU hardware. 

A second challenge arising when modeling driver behavior is the difficulty of acquiring annotated data. Driver behavior annotation techniques can include driver surveys \cite{Chhabra2019a, Hong2014} or manual annotation by domain experts \cite{mundke2006}. 
These techniques can be both time consuming and expensive, since they require human annotators. Therefore, it is beneficial to be more careful when selecting the data to be used for training the models. The idea of \textit{active learning} is to only present the most informative samples to the human annotator and thus minimize the amount of expensive annotation. Active learning is therefore an increasingly popular technique in situations where only a small subset of a large pool of unlabeled data can feasibly be annotated \cite{Settles2008a, tong2001active}. 

In this paper, we investigate several methods to mitigate these two challenges. Our contribution to address the first challenge, a problem setting which we refer to as \textit{passive learning} as a contrast to the second challenge, is to investigate two non-recurrent models for modeling driving behavior: \textit{self-attention} (SA) models and \textit{convolutional neural networks} (CNN) with \textit{joint recurrence plots} (JRP) created from driving signals. We compare them with recurrent models such as LSTM. 

To address the second challenge, we employ \textit{active learning} in order to mitigate the lack of annotated data. The active learning methods evaluated in this study are: uncertainty sampling, query by committee (QBC), and active deep dropout (ADD).

\section{BACKGROUND  \&  METHODS}
\label{sec:background}

Driving behavior has been studied over the last decade with various models and features. Some modeling methods investigated in literature include Gaussian Mixture Models (GMM) \cite{Nishiwaki2007a,Nishiwaki2007b,Choi2007}
, Hidden Markov Models (HMM) \cite{Choi2007,Kuge2000ADB,Takano2008RecognitionOH}
, and neural networks \cite{Shahverdy2020a,MacAdam1998a,Zhang2019b}. 
Literature on driver behavior can also differ in the type of features used for modeling. Examples of such features are: Controller Area Network (CAN) bus signals \cite{Choi2007,Zhang2019b,Liu2017a,Zhang2016a}
, smartphone sensors \cite{Hong2014,Zhang2016a,Junior2017a}
, questionnaires \cite{Chhabra2019a, Ishibashi2007}
, and video feeds \cite{Zhang2018a,Ihme2018RecognizingFO}.  

To determine the utility of each of the suggested features, a more comprehensive overview of the problem is useful. Elamrani Abou Elassad \textit{et al.} \cite{ElamraniAbouElassad2020a} introduce a conceptual framework which decomposes the term \textit{driving behavior} into three types of phenomena: \textit{driving events} (i.e., driving operations performed by the driver, such as tailgating, turning, etc.), the physiological state of the driver, and the psychological state of the driver. These phenomena can be modeled using various metrics, including CAN-bus signals, camera feeds focusing on drivers, and electrocardiograms (used to measure heart rate).

Below, we briefly outline the topics we investigate in this study. Sections \ref{sec:background_driver_behavior} through \ref{sec:background_recurrence_plots} relate to the passive learning setting, while Section \ref{sec:background_active_learning} outline the active learning approaches we consider.

\subsection{Driver behavior and energy consumption}
\label{sec:background_driver_behavior}
A number of past research studies have explored how driver behavior affects energy or fuel consumption. Younes \textit{et al.} \cite{Younes2013a} analyzed factors affecting energy consumption in electric vehicles, including driver behavior. They provide qualitative definitions for calm, normal, and aggressive driving styles, which we have adopted in our study. They also compute parameters that were found to have a correlation with driving behavior, which we employ in our annotation system and integrate within our driver behavior model. 

We also quantify aggressiveness through the percentage of time the driver adopts such behavior, as proposed by Constantinescu \textit{et al.} \cite{Constantinescu2010a}, and through the distance to the vehicle in the front, as proposed by MacAdam \textit{et al.} \cite{MacAdam1998a}.

\subsection{Self-attention models}
\label{sec:background_self_attention}

The first of the two non-recurrent models we investigate in our study is the self-attention model, which was introduced  by Vaswani \textit{et al.} \cite{Vaswani2017b}. The authors argue that there is a paradigm shift towards attention-based models and away from recurrent models like RNN, LSTM, and GRU. The reasoning behind this claim is based on three advantageous properties of self-attention: lower computational complexity per network layer, better capabilities for parallelized computations, and a higher ability to capture long-range dependencies in sequential data.

The authors state that the computational complexity per layer for self-attention models is $O(n^2 \cdot d)$, where $n$ is the sequence length and $d$ is the number of dimensions. They argue that this is beneficial for natural language processing (NLP), since the sequence (i.e., sentence) length in language tasks is usually much lower than the number of dimensions (e.g., the amount of words in an English dictionary). Although the self-attention model was originally used for NLP tasks, Mahmud \textit{et al.} \cite{Mahmud2020} utilized the idea for time-series data by including self-attention blocks in their architecture for Human Activity Recognition (HAR). In their work, sensor values and time windows correspond to words and sentences in the NLP setting, respectively.

The mechanism of self-attention can be described in the following way. Let $\mathbf{X} \in \mathbb{R}^{n \times d}$ be the matrix containing the inputs, where $d$ is the number of dimensions of the input and $n$ is the sequence length. Furthermore, let $\mathbf{W}_Q, \mathbf{W}_K, \mathbf{W}_V \in \mathbb{R}^{d \times d}$ be learnable weight matrices. These learnable weights are multiplied by the inputs to create the following matrices:
\begin{equation}
    \begin{split}
        \mathbf{Q}^T = \mathbf{W}_Q \cdot \mathbf{X}^T \\
        \mathbf{K}^T = \mathbf{W}_K \cdot \mathbf{X}^T \\
        \mathbf{V}^T = \mathbf{W}_V \cdot \mathbf{X}^T .
    \end{split}
    \label{eq: query, keys and values}
\end{equation}

Matrices $\mathbf{Q}, \mathbf{K},$ and $\mathbf{V}$ are known as \emph{query}, \emph{key}, and \emph{value} matrices respectively. They are used to weigh the relationships between different positions in the input sequence:
\begin{equation} \label{eq:attention}
    \text{Attention}(\mathbf{Q}, \mathbf{K}, \mathbf{V}) = \text{softmax}\Big(\frac{\mathbf{Q} \cdot \mathbf{K}^{T}}{\sqrt{d}}\Big) \cdot \mathbf{V} .
\end{equation}

Vaswani \textit{et al.} proposed the scaling factor of $1 / \sqrt{d}$ in order to prevent the gradients from vanishing for large values of $d$. This results in what the authors refer to as \textit{scaled dot-product attention}.

Note, however, that by using the method above, the positional information of the input sequences is lost. To address this issue, Vaswani \textit{et al.} introduced a positional encoding method to account for positional information when feeding each input through the model. A positional vector $\mathbf{p}_t$ is computed from the sequence position, and later added to the input.

Let \textit{t} be the position of the input in a given sequence (e.g., a time window of sensor values) and let \textit{d} be the number of input dimensions and \textit{i} be a dimension index. Also, let $j = \lceil{\frac{i}{2}} \rceil$. Then each element $p_{t,i}$ of the positional vector $\mathbf{p}_t$ is computed as follows:
\begin{equation}
p_{t,i} = 
  \begin{cases}
      \sin( \frac{t}{10000^{2j / d}}),  & \text{if \textit{i} is even} \\
      \cos( \frac{t}{10000^{2j / d}}),  & \text{if \textit{i} is odd}
  \end{cases} .
\end{equation}

\subsection{CNN with recurrence plots}
\label{sec:background_recurrence_plots}

Signals can be processed into \textit{recurrence plots} \cite{eckmann1987recurrence}. These plots visually show if a signal at time \textit{i} is similar to the signal at time \textit{j}. A recurrence plot is defined as a matrix $\mathbf{R}$, where each element is computed as follows:
\begin{equation}
    \mathbf{R}_{i,j}= 
    \begin{cases}
        1,& \text{if } || \mathbf{x}_i -\mathbf{x}_j  || \leq \epsilon \\
        0,              & \text{otherwise,}
    \end{cases}
    \label{eq: recurrence plots}
\end{equation}
where $\mathbf{x}_i$ is a signal and $\epsilon$ is a threshold that controls the permitted difference between $\mathbf{x}_i$ and $\mathbf{x}_j$. 

Shahverdy \textit{et al.} \cite{Shahverdy2020a} pre-processed recorded CAN-bus signals (e.g., throttle and speed) into recurrence plots and used the plots to train CNN models for driver behavior identification. The authors classify driving behavior into five categories: distracted, drowsy, drunk, aggressive, and safe. They argue that the spatial information in the recurrence plots has benefits beyond capturing temporal dependencies in the signals. As previously mentioned, one additional benefit is the ability to avoid sequential computations. Our study aims to investigate these properties of recurrence plots. We also extend the methods of Shahverdy \textit{et al.} by investigating the possibility of converting several driving signals into one \textit{joint recurrence plot}.

A \textit{joint recurrence plot} (JRP) is a method for combining several recurrence plots into a single plot, using the \textit{Hadamard product}. For matrices $\mathbf{A}$, $\mathbf{B}$ and $\mathbf{C}$ of equal dimension, it is defined as an element-wise multiplication of matrices $\mathbf{A}$ and $\mathbf{B}$, such that $\mathbf{C}_{i,j}= (\mathbf{A})_{i,j}(\mathbf{B})_{i,j}$.

\subsection{Active learning}
\label{sec:background_active_learning}
Active learning is a class of methods to select and annotate only the most informative samples for the training set. Examples of applications where this paradigm is useful include tasks in which annotated data is expensive or scarce, such as text classification \cite{lewis}, speech recognition \cite{Gokhan}, troubleshooting \cite{ChenRCK17}, driving trajectory analysis \cite{JARL2022104972}, cancer diagnosis \cite{dLiuYing} and medicine \cite{AL_drug_discovery}. Conventional active learning methods include \emph{uncertainty sampling} and \emph{query by committee} \cite{Settles2010b}.  
The former method queries the most uncertain samples according to a specified informativeness measure. The latter queries the most informative samples by using a committee of models. Both methods present informative samples to a human annotator whose role is to annotate the samples. 

We outline the informativeness measures we use with uncertainty sampling. \textit{Least confidence} is the simplest measure, and involves choosing the instance $\mathbf{x}$ in which the model is least confident. The queried instance can be formulated as follows:
\begin{equation}
    \text{arg}\max_{\mathbf{x}} 1 - P_{\theta}(\hat{y} | \mathbf{x}) ,
\end{equation}
where $\hat{y} = \text{arg}\max_y P_{\theta}(y | \mathbf{x})$ represents the predicted class label for input $\mathbf{x}$ according to the highest posterior probability for the model $\theta$ \cite{Settles2010b}. This informativeness measure only takes into account the class in which the model is least confident, and therefore does not distinguish well between classes in multi-class classification. 

The following measure takes into account the top two classes:
\begin{equation}
    \text{arg}\min_{\mathbf{x}} P_{\theta}(\hat{y}_1 | \mathbf{x}) - P_{\theta}(\hat{y}_2 | \mathbf{x}) ,
\end{equation}
where $\hat{y}_1$ is the most probable class and $\hat{y}_2$ is the second most probable class under the model $\theta$. This informativeness measure is known as \textit{margin}. It is better suited than the least confidence measure for distinguishing classes in multi-class problems, and is studied in detail for deep learning in \cite{Bossr2020d}. 

However, both the margin and least confidence measures might be unsuited for problems where there is a large amount of class labels, since they ignore information about most of the class labels. The following informativeness measure takes into account all class labels:
\begin{equation}
    \text{arg}\max_{\mathbf{x}} - \sum_i P_{\theta}(y_i | \mathbf{x})\text{log}P_{\theta}(y_i | \mathbf{x}) .
\end{equation}

This measure is commonly referred to as \textit{entropy} and is the most popular informativeness measure used within active learning \cite{Settles2010b}. Informally, it can be described as choosing the instance $\mathbf{x}$ which has the lowest predictability amongst all classes (and, equivalently, the highest entropy).

In query by committee, the level of disagreement between committee members is instead typically assessed through the \textit{vote entropy} or \textit{Kullback-Leibler divergence} informativeness measures \cite{Settles2010b}.
The former is a variation of the entropy informativeness measure, taking into account the number $V(y_i)$ of committee members \textit{voting} for each label $y_i$ (i.e., predicting $y_i$ as the most probable label) over the size of the committee $N$:
\begin{equation} \label{eq:voteen}
    \text{arg}\max_{\mathbf{x}} - \sum_i \frac{V(y_i)}{N}\text{log}\frac{V(y_i)}{N} .
\end{equation}

The latter measure employs the Kullback-Leibler divergence, and therefore considers the most informative queries to be the ones where the members' label distribution differs the most from the consensus, as described in Eq. \ref{eq:kldiv}, where $\theta^{(j)}$ represents a single member of the committee, and $C$ represents the whole committee:
\begin{equation}\label{eq:kldiv}
    \text{arg}\max_{\mathbf{x}}\frac{1}{N}\sum_{j=1}^ND(P_{\theta^{(j)}}||P_C) .
\end{equation}

The Kullback-Leibler divergence is defined as:
\begin{equation}
    D(P_{\theta^{(j)}}||P_C) \coloneqq \sum_iP_{\theta^{(j)}}(y_i|\mathbf{x})\text{log}\frac{P_{\theta^{(j)}}(y_i|\mathbf{x})}{P_C(y_i|\mathbf{x})}
\end{equation}
and the consensus probability of $y_i$ being the correct label is defined as:
\begin{equation}
    P_C(y_i|\mathbf{x}) \coloneqq \frac{1}{N}\sum_{j=1}^NP_{\theta^{(j)}}(y_i|\mathbf{x}) .
\end{equation}

Finally, Gammelsæter \cite{Gammelsaeter2015b} proposed a novel query by committee approach called \emph{active deep dropout} (ADD). The idea behind this technique is to implement a committee of models through dropout regularization. 

\section{METHODOLOGY}
This study investigates driver behavior classification on two naturalistic driving datasets: the data collected on a test track by the authors of this study, and the data collected (with permission and pseudonymized) from real drivers using connected test vehicles over several months. The latter dataset is annotated according to a set of rules and parameters that quantifies the aggressiveness of the driving style.

\subsection{Test track dataset}
In this dataset, two drivers have emulated aggressive, normal, and cautious driving styles while driving a battery electric vehicle on a test track. Each style was used for an approximately equal number of laps and was maintained throughout the whole lap. The drivers emulated the driving styles using a set of qualitative instructions from previous literature and domain experts. The aggressive driving style was emulated by performing fast accelerations and decelerations, by changing lanes often and abruptly, and by driving close to the speed limit. The normal style was emulated by performing smoother accelerations and decelerations, by changing lanes gradually and only if necessary, and by keeping a speed slightly below the speed limit. The cautious style was emulated like the normal style, but with very careful accelerations and decelerations, and a velocity significantly lower than the speed limit.

\subsection{Real drivers dataset}
The second dataset was collected from a set of vehicles driven by real drivers in a naturalistic way, and it consists of a combination of sensors and navigation map data, such as the longitudinal and lateral speed and acceleration, the current and next speed limit, and the road gradient. The road type affects driver behavior \cite{ElamraniAbouElassad2020a, Younes2013a}, which we control by identifying and using a frequent commuting route.

\subsection{Window segmentation}
We use sliding window segmentation to separate the driving data into time windows: for the majority of the experiments, we extract non-overlapping windows of either $5$ or $10$ seconds. The chosen window sizes have been used in previous work \cite{Peng2017} and are long enough to include complete driving manoeuvres, such as abrupt accelerations or lane changes, while also presenting a lower risk than larger window sizes of including more than one driving style. The annotation and classification processes are applied to individual windows. For the classification tasks, we perform stratified $5$-fold cross validation and set the same random seed across experiments. Correspondingly, we define the same set of seeds for experiments with active learning. 

\subsection{Annotation}
The windows are annotated using a majority-class system of driving parameters and rules found in accordance with domain experts and in previous work \cite{Constantinescu2010a}. Table \ref{table:rules} displays the set of rules and the corresponding classes adopted in the annotation process: the rules take into consideration speeding behavior and time gap, i.e., the difference in time between two adjacent vehicles.

Other parameters considered for the annotation were previously proposed by Younes \textit{et al.} \cite{Younes2013a}, and were found by the authors to correlate with driver aggressiveness: positive kinetic energy (PKE), relative positive acceleration (RPA), root mean square of the power factor (RMSPF), and mean and standard deviation of jerk (change in acceleration). PKE is a measure of the intensity of positive acceleration manoeuvres, defined here as:
\begin{equation} \label{eq:pke}
    PKE \coloneqq \frac{\sum_i(v^2_{i+1} - v^2_i)}{D}, v_{i+1} > v_i ,
\end{equation}
where $v_i$ is the vehicle's speed at time step $i$ and $D$ is the total trip distance (in our case, the total window distance). RPA is defined as:
\begin{equation} \label{eq:rpa}
    RPA \coloneqq \frac{\sum_i(v_i * a_i^+)}{D} ,
\end{equation}
where $a^+_i$ is the vehicle's positive acceleration at time step $i$. RMSPF is defined as:
\begin{equation}
    RMSPF \coloneqq \sqrt{\frac{1}{n}\sum_{i=1}^n(2*v_i*a_i)^2} ,
\end{equation}
where $2*v_i*a_i$ is the power factor at time step $i$. These five parameters (PKE, RPA, RMSPF, jerk mean and jerk standard deviation) are computed for $10$-second windows of the data generated on the test track, and their Pearson coefficient is calculated to verify if they correlate with driving style. We find that they all present significant correlation (average $r$=$0.32$, \textit{p}-values $< 0.001$).

In a second phase, the probability density function for the parameters of Younes \textit{et al.} is estimated through Gaussian kernel density estimation, on the samples computed using the windows of the test track data. We also compute the same parameters for each window extracted from the unlabeled data collected from real drivers, and the probability of the feature value belonging to each class is inferred from the density function. The class with the highest probability for that value is treated as its label. Finally, each window is annotated with the class that agrees the most with the rules in Table \ref{table:rules} and the probabilities of the five driving parameters of Younes \textit{et al.}

\begin{table}[h!]
\centering
\caption{Rules for annotation based on speed and time gap}
\begin{tabular}{ll}
\toprule[1pt]\midrule[0.3pt]
\multicolumn{1}{c}{\textbf{Rule}} & \multicolumn{1}{c}{\textbf{Class}} \\ \hline
\begin{tabular}[c]{@{}l@{}}\textbf{Speeding}: driving at least\\ 5 km/h above the speed limit,\\ for at least 20\% of the time\end{tabular} & \begin{tabular}[c]{@{}l@{}}If true and \textbf{slow driving} is false:\\ \textbf{aggressive}\\ If true and \textbf{slow driving} is true:\\ \textbf{normal}\end{tabular} \\ \hline
\begin{tabular}[c]{@{}l@{}}\textbf{Slow driving}: driving at least\\ 5 km/h below the speed limit,\\ for at least 10\% of the time,\\ when the vehicle in front\\ is at least $20$ meters away\end{tabular} & \begin{tabular}[c]{@{}l@{}}If true and \textbf{speeding} is false:\\                  \textbf{cautious}\\ If true and \textbf{speeding} is true:\\ \textbf{normal}\end{tabular} \\ \hline
\begin{tabular}[c]{@{}l@{}}\textbf{Low time gap}: when the\\ time gap is at most 1 second\\ from the vehicle in front,\\ for at least 20\% of the time\end{tabular} & \begin{tabular}[c]{@{}l@{}}If true and \textbf{high time gap} is false:\\ \textbf{aggressive}\\ If true and \textbf{high time gap} is true:\\                   \textbf{normal}\end{tabular} \\ \hline
\begin{tabular}[c]{@{}l@{}}\textbf{High time gap}: when the\\ time gap is at least 2.5 seconds\\ from the vehicle in front,\\ for at least 10\% of the time,\\ whenever that vehicle is\\ closer than 50 meters away\end{tabular} & \begin{tabular}[c]{@{}l@{}}If true and \textbf{low time gap} is false:\\ \textbf{cautious}\\ If true and \textbf{low time gap} is true:\\ \textbf{normal}\end{tabular} \\ \midrule[0.3pt]\bottomrule[1pt]
\end{tabular}
\label{table:rules}
\end{table}

\subsection{Driving features}
When training the neural network models, we select the driving features that depend the most on the actions of the driver (and the speed limit), for a total of $8$ features: longitudinal acceleration, speed, speed limit, percentage of pressure on the acceleration pedal, lateral acceleration, steering wheel angle, rotational speed of the steering wheel, and distance from the vehicle in the front. The features are scaled through standardization.

\subsection{Passive learning}\label{sec:methodology_passive_learning}
This study considers the comparison of four models with different architectures: a one-dimensional convolutional neural network (1D-CNN), a self-attention (SA) network, a CNN with joint recurrence plots (JRP), and a long short-term memory (LSTM) recurrent neural network. The first three models are considered to be non-recurrent while the last one has a recurrent architecture.

The layer setup of the 1D-CNN consists of two convolutional layers with filter width equal to the number of time steps in the windows of data. Batch normalization and dropout are added to stabilize the training process and prevent overfitting, respectively. The LSTM architecture consists of a unidirectional LSTM layer followed by a fully connected layer and by batch normalization and dropout. 

Two different versions of the self-attention model are implemented: one following the architecture originally proposed by Vaswani \textit{et al.} \cite{Vaswani2017b}, and the other according to the architecture proposed by Mahmud \textit{et al.} \cite{Mahmud2020}, which is optimized for time-series data. The first is used with the test track data, while the second is used with the data collected from real drivers, as each model is observed to perform slightly better than the other on their respective datasets. Both models are trained with the adaptive moment estimation optimizer and with early stopping. 

\subsection{Active learning}
A cumulative training approach is implemented according to the procedure described by Bossér \textit{et al.} \cite{Bossr2020d}: first, the classifier is trained on the labeled set $\mathcal{L}$; second, the unlabeled samples $\mathcal{U}$ are picked uniformly at random (for random sampling, corresponding to the passive learning setting) or ranked according to the chosen informativeness measure (for active learning methods), and batch $\mathcal{B} \subseteq \mathcal{U}$ of the $n$ top samples is selected; finally, the classifier is re-trained on $\mathcal{L} \cup \mathcal{B}$ after parameter re-initialization. This process is repeated until $|\mathcal{L} \cup \mathcal{B}|$ is $80\%$ of the size of the whole dataset. All experiments are carried out on non-overlapping windows of $10$ seconds on the real driver data. The size of the test dataset is set to $20\%$, and it is fixed throughout all iterations of the same experiment.

We investigate and explore two different active learning approaches: uncertainty sampling and query by committee. For uncertainty sampling, we compare the least confidence, margin, and entropy informativeness measures. For query by committee, we consider two methods: standard query by committee, and active deep dropout. 

For standard query by committee, we employ a committee of three members: a 1D-CNN, an LSTM, and a CNN-LSTM. The architectures of the 1D-CNN and LSTM committee members are described in Section \ref{sec:methodology_passive_learning}. The CNN-LSTM architecture consists of two convolutional layers (as in the 1D-CNN), followed by two unidirectional LSTM layers, a fully connected layer and a dropout layer. 

The active deep dropout method consists of training a \textit{parent} model and then generating a committee of $5$ members, where each member has the same architecture as the parent and a different dropout configuration. The committee members perform inference on the unlabeled set through a single forward pass per iteration. Both of the query by committee methods employ the vote entropy and Kullback-Leibler divergence informativeness measures. In all active learning experiments, we first train the models on $10\%$ of randomly picked data and then on $5\%$ increments of the samples selected through the chosen informativeness measure. We then assess the test accuracy for $14$ iterations, i.e., starting from $15\%$ up to $80\%$ of the training data.

\section{RESULTS \& DISCUSSION}

\subsection{Passive learning}

\begin{table}[t!]
\centering
\caption{Models' performance on the test track data}
\begin{tabular}{lcccc}
\toprule[1pt]\midrule[0.3pt]
\textbf{Model}  & \textbf{\begin{tabular}[c]{@{}c@{}}Accuracy\\     \end{tabular}}                                                         & \textbf{\begin{tabular}[c]{@{}c@{}}Weighted-avg\\ Precision\end{tabular}}   & \textbf{\begin{tabular}[c]{@{}c@{}}Weighted-avg\\ Recall\end{tabular}}& \textbf{\begin{tabular}[c]{@{}c@{}}AUC\end{tabular}} \\ 
\hline
& & $5$ seconds & &\\

\hline
\textbf{LSTM} & \textbf{0.82} & 0.81 & 0.78 & \textbf{0.92} \\ 
\textbf{SA} & 0.80 & 0.78 & 0.78 & 0.91 \\ 
\textbf{1D-CNN} & 0.76 & 0.73 & 0.75 & 0.90 \\  
\textbf{JRP} & 0.46 & 0.50 & 0.50 & 0.70 \\  

\hline
& & $10$ seconds & &\\

\hline
\textbf{LSTM} & \textbf{0.87} & 0.88 & 0.87 & \textbf{0.95} \\  
\textbf{SA} & 0.81 & 0.83 & 0.81 & 0.93 \\  
\textbf{1D-CNN} & 0.79 & 0.86 & 0.80 & 0.93 \\  
\textbf{JRP} & 0.40 & 0.28 & 0.36 & 0.60 \\  

\hline
& & \begin{tabular}[c]{@{}c@{}}5 seconds\\ $50\%$ overlap\end{tabular}  &\\

\hline
\textbf{LSTM} & \textbf{0.91} & 0.92 & 0.90 & 0.96 \\ 
\textbf{SA} & 0.84 & 0.83 & 0.83 & 0.94 \\  
\textbf{1D-CNN} & 0.88 & 0.91 & 0.87 & \textbf{0.97} \\  
\textbf{JRP} & 0.49 & 0.49 & 0.50 & 0.70 \\ 
\midrule[0.3pt]\bottomrule[1pt]

\end{tabular}
\label{tab:results_test_track}
\end{table}

\begin{table}[t!]
\centering
\caption{Models' performance on the classes}
\begin{tabular}{lcccc}
\toprule[1pt]\midrule[0.3pt]
\textbf{Model} & \textbf{Precision} & \textbf{Recall} & \textbf{F1 score}  \\
\hline 
LSTM & & & &\\
\hline
\textbf{Aggressive} & 1.00 & 1.00 & 1.00 \\
\textbf{Normal} & 0.59 & 0.68 & 0.63 \\ 
\textbf{Cautious} & 0.77 & 0.69 & 0.73 \\ 
\hline 
Self-attention & & & &\\
\hline
\textbf{Aggressive} &0.96 & 0.96 & 0.96 \\ 
\textbf{Normal} & 0.75 & 0.53 & 0.62 \\ 
\textbf{Cautious} & 0.73 & 0.87 & 0.79 \\ 
\hline 
1D-CNN & & & &\\
\hline
\textbf{Aggressive} & 1.00 & 0.96 & 0.98 \\ 
\textbf{Normal} & 0.51 & 0.68 & 0.58 \\ 
\textbf{Cautious} & 0.74 & 0.62 & 0.67 \\ 
\hline 
JRP & & & &\\
\hline
\textbf{Aggressive} & 0.50 & 0.20 & 0.29 \\ 
\textbf{Normal} & 0.50 & 0.75 & 0.60 \\
\textbf{Cautious} & 0.50 & 0.10 & 0.67 \\ 
\midrule[0.3pt]\bottomrule[1pt]

\end{tabular}
\label{tab:results_classes}
\end{table}

Table \ref{tab:results_test_track} shows the performance of the models on the test track data, with three different window configurations: 5s windows, 10s windows, and 5s windows with 50\% overlap. The different window configurations show that the model performance increases with the window overlapping technique. This improvement suggests that more training data can potentially improve the models even further. 

Improvement was also observed whenever 10s windows were used instead of 5s windows. This may suggest that longer time windows have the potential to improve the performance of the models, by including more driving characteristics present in the signals. Studies that also use window sizes of 5s and 10s include \cite{Dai2010} and \cite{MacAdam1998a}. Zhang \textit{et al.} \cite{Zhang2019b} used longer time windows of up to 1 minute. However, the authors attempted to model the specific driving styles of individual drivers (also known as driver identification). Studies on driver identification may use longer time windows in order to clearly differentiate the driving style of each driver. 

Vaswani \textit{et al.} \cite{Vaswani2017b} argue that the self-attention layer is better suited for modeling longer time sequences. They propose that self-attention layers are more beneficial since the ``maximum path length between any two input and output position'' is shorter. This property is also prevalent in CNN architectures, and allows the self-attention model to capture long-term dependencies better than recurrent models, such as LSTM. 

However, our chosen window lengths of 5s and 10s did not seem to be detrimental to the performance of the LSTM model. In fact, LSTM outperformed all of the non-recurrent models, for all performance metrics (Table \ref{tab:results_test_track}). This suggests that, for our window lengths, LSTM is still the best model architecture. However, LSTM might face challenges if it is used to model driver behavior in longer time sequences. 

\begin{table}[H]
\centering
\caption{Computation Time}\label{tab:results_training_time}
\begin{tabular}{lcccc}
\toprule[1pt]\midrule[0.3pt]
\textbf{Model} &  \textbf{\begin{tabular}[c]{@{}c@{}}5s\\ windows\end{tabular}} & \textbf{\begin{tabular}[c]{@{}c@{}} 10s \\ windows\end{tabular}} &  \textbf{\begin{tabular}[c]{@{}c@{}} 50s \\ windows\end{tabular}} \\ 

\hline
\textbf{1D-CNN} & 1ms & 2ms & 6ms \\ 
\textbf{JRP} & 3ms & 15ms & 32ms \\ 
\textbf{LSTM} & 6ms & 15ms & 42ms \\ 
\textbf{Self-attention} & 46ms & 61ms & 424ms \\ 

\midrule[0.3pt]\bottomrule[1pt]

\end{tabular}
\end{table}

In terms of computational efficiency, the differences between the LSTM and the non-recurrent models are shown in Table \ref{tab:results_training_time}. The value shown in each cell is the time it takes for the model to perform forward propagation of a batch of 5 samples. All models have the same number of trainable parameters (i.e., 4700) in order to fairly compare the propagation time of each model.

Most of the non-recurrent models outperformed the LSTM model, except the model using the self-attention architecture. These results indicate that the self-attention model has the worst computational performance. A possible reason for this is that the computational complexity per layer of the self-attention architecture is, as stated in Section \ref{sec:background_self_attention}, $O(n^2 \cdot d)$. Our datasets contain longer sequences (with 50 or 100 time steps per window) than the number of sensors (12 dimensions, including throttle, steering, etc.). The parallelization property of self-attention architectures did not seem to improve the computational performance.

Some concluding remarks can be stated regarding our investigation of non-recurrent models. The model using self-attention architecture can successfully detect the aggressive class (Table \ref{tab:results_classes}), which we consider the most interesting case, due to its correlation with fuel consumption. However, the parallelization property of the self-attention model does not provide benefits in terms of computational speed, for our window configurations and window processing techniques. When it comes to JRP, the model did not perform well overall, with particularly low precision and recall for the aggressive class. However, in terms of computational efficiency, this architecture outperformed LSTM.

\subsection{Active learning}
\begin{figure}[h!]
     \centering
     \subfigure[1D-CNN]{        \includegraphics[width=0.75\columnwidth]{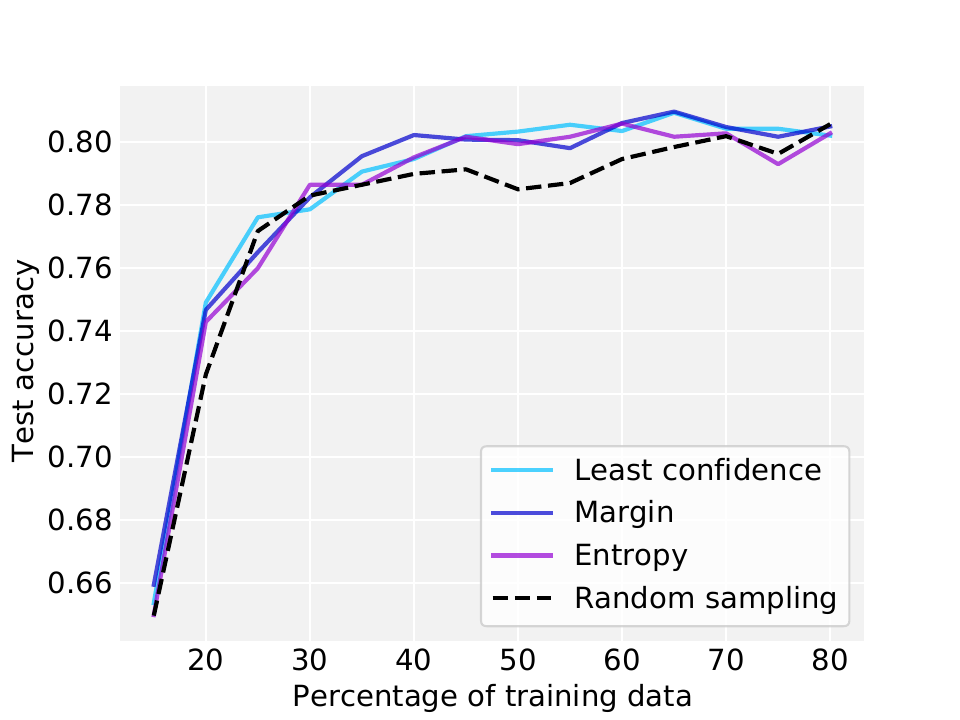}\label{fig:us_cnn}}
     \\
     \subfigure[LSTM]{\includegraphics[width=0.75\columnwidth]{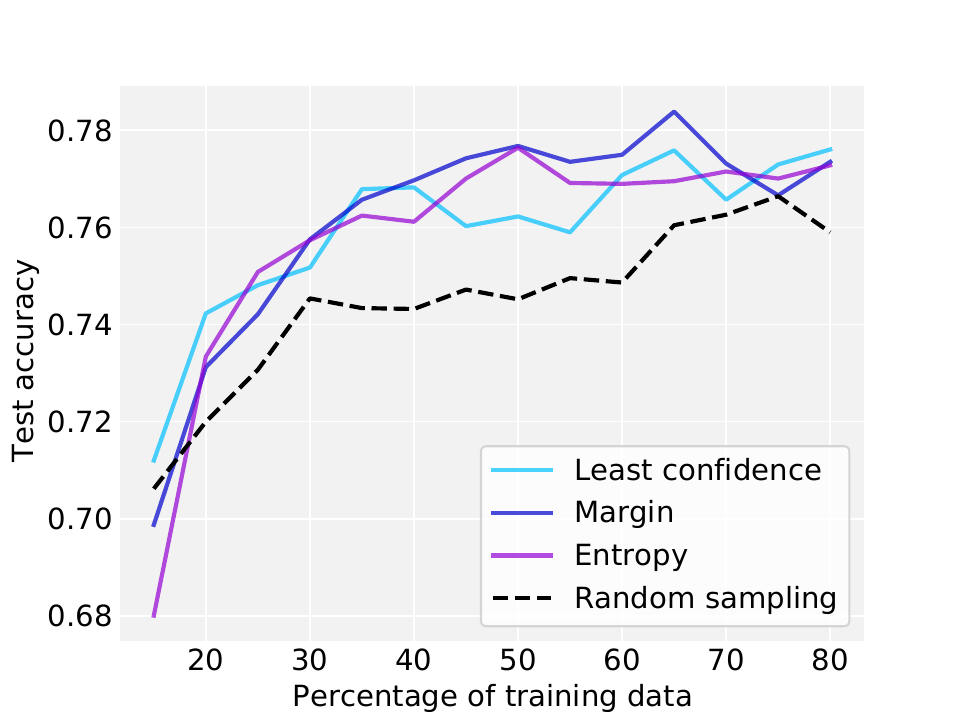}\label{fig:us_lstm}}
     \caption{Comparison of uncertainty sampling methods with random sampling on the (a) 1D-CNN and (b) LSTM models.}
\label{fig: us}
\end{figure}

We evaluate the effectiveness of the active learning methods through learning curves, each showing the test accuracy of models trained on different percentages of actively acquired training data. Fig. \ref{fig: us} shows the learning curves of the uncertainty sampling experiments on (a) the 1D-CNN model and (b) the LSTM model. The uncertainty sampling techniques perform reliably better than random sampling, except for the early iterations with the margin and entropy informativeness measures for the LSTM model. The improvement over random sampling is particularly evident in this model.

\begin{figure}[h!]
     \centering
     \subfigure[1D-CNN]{        \includegraphics[width=0.75\columnwidth]{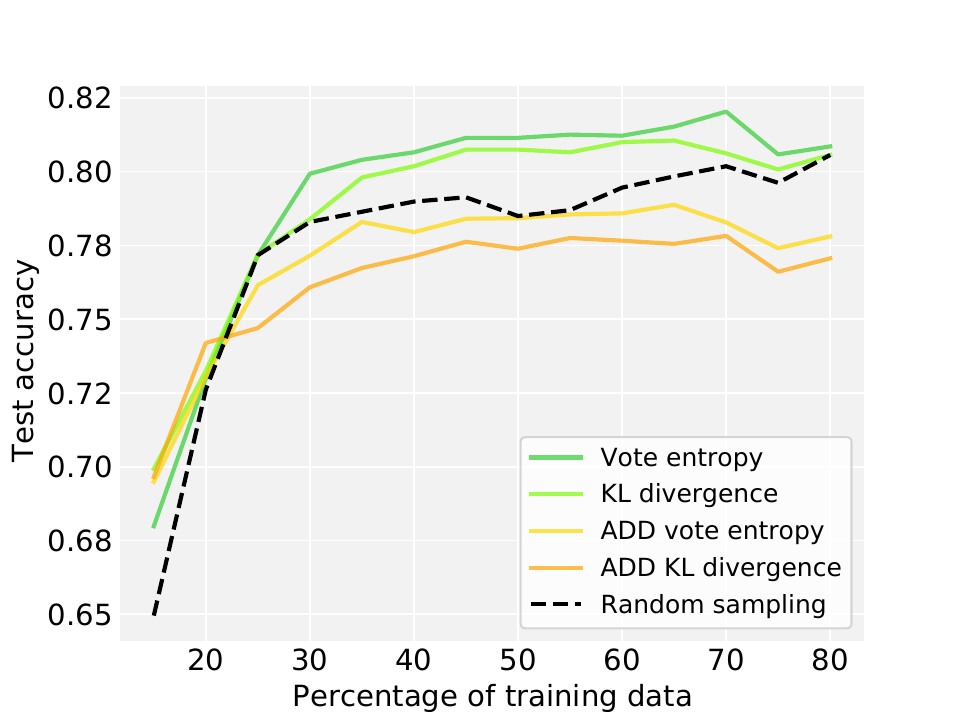}\label{fig:qbc_cnn}}
     \\
     \subfigure[LSTM]{\includegraphics[width=0.75\columnwidth]{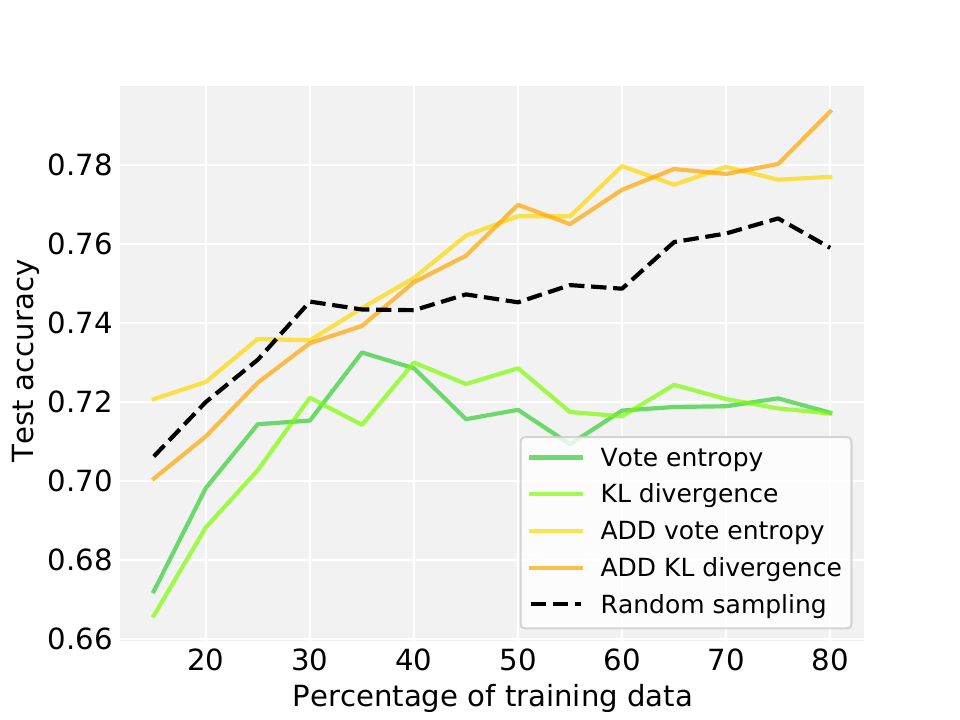}\label{fig:qbc_lstm}}
     \caption{Comparison of query by committee methods with random sampling on the (a) 1D-CNN and (b) LSTM models.}
\label{fig: qbc}
\end{figure}

Fig. \ref{fig: qbc} shows the corresponding learning curves of the query by committee experiments on the same models. The parent model for active deep dropout is 1D-CNN in Fig. \ref{fig:qbc_cnn} and LSTM in Fig. \ref{fig:qbc_lstm}. The results present a higher variability than the uncertainty sampling methods: regular query by committee is successful for the 1D-CNN model but not for the LSTM model, whereas the opposite can be observed for active deep dropout methods. 

We speculate that the predictions of the LSTM model are helpful for the 1D-CNN model, and that active learning on the 1D-CNN model consequently yields better results when the LSTM model is included as a committee member. As a contrast, we speculate that the predictions of the 1D-CNN model are not as helpful for the LSTM model. This is consistent with the findings of Lowell \textit{et al.} \cite{chall_al2019}, who found a strong coupling between acquired training sets and the model with which they were acquired. Moreover, active learning methods imply a bias in sampling, i.e., the violation of the assumption that the training samples are independent and identically distributed, and sampled from the population distribution \cite{chall_al2019, farquhar_statistical_2020}. This might explain the variability in performance of the models trained on actively sampled data.

\section{CONCLUSION \& FUTURE WORK}

In this paper, we investigated recurrent and non-recurrent neural network architectures for modeling driver behavior, in passive and active learning settings. The two non-recurrent models investigated for the passive learning challenge exhibited mixed results. Starting with self-attention, this model displayed good performance in terms of precision and recall (especially for the aggressive driving behavior class) when compared to the other models. However, the self-attention model also exhibited the worst performance when it comes to computational speed. This result most likely stems from the fact that self-attention layers are more computationally expensive than recurrent layers, whenever the sequence length (i.e. window length) is significantly larger than the dimension (i.e. number of sensors). This property seems to have had more impact on the computational speed of the model than the parallelization property of the self-attention architecture. A future research suggestion is to apply the self-attention architecture on longer time windows, e.g., for driver identification \cite{Zhang2019b}. The idea is then to not only test its ability to utilize parallelization, but also the ability to capture long-term dependencies.

In contrast, the other non-recurrent model, JRP with CNN, may have benefited from the parallelization property in terms of computational speed. The computation time of the model was the second best, only outperformed by the simpler 1D-CNN model. However, the model did not perform well w.r.t. any performance metric. Our observation is that this model quickly overfits. Results did, however, improve whenever the dataset was pre-processed with some overlap of the windows. This suggests that more training data could improve the model. Future research can be to investigate different windowing techniques to improve the performance of the JRP models.

Our final investigation involved active learning. The experiments showed that uncertainty sampling methods brought an improvement in test accuracy over training with passive learning, and therefore reduced the need for training data. Other active learning techniques, such as query by committee methods, in contrast, showed more variable results. We hypothesize that this variability is due to the bias introduced by non-random sampling, and also due to the coupling between actively sampled datasets and trained model. In this study, we expanded on Gammelsæter's work by applying the proposed active deep dropout technique on a new dataset. For future work, it may be interesting to expand further on active deep dropout with different committee setups. It may also be interesting to explore different regularization techniques to create committees, and to incorporate active learning informativeness measures that are more suited to time-series data, such as those proposed in \cite{Peng2017}. To deal with lack of sufficient labeled data, an alternative approach would be generating synthetic data (similar to the driving scenarios generated by Generative Adversarial Networks in \cite{abs-2007-14524,DemetriouARC20}) which we postpone to future work.





\section*{ACKNOWLEDGMENT}

Part of this study was performed by Federica Comuni and Christopher Mészáros as a master thesis project with Volvo Car Corporation, which also provided support, data and resources for the study. Niklas Åkerblom is a PhD student employed by Volvo Car Corporation, with funding by the Strategic Vehicle Research and Innovation Programme (FFI) of Sweden, through the project EENE (reference number: 2018-01937).


\bibliographystyle{ieeetr}
\bibliography{bibliography}

\end{document}